\documentclass[letterpaper]{article} 
\usepackage{aaai2026}  
\usepackage{times}  
\usepackage{helvet}  
\usepackage{courier}  
\usepackage[hyphens]{url}  
\usepackage{graphicx} 
\usepackage{natbib}  
\usepackage{caption} 
\usepackage{algorithm}
\usepackage{algorithmic}

\usepackage{latexsym}
\usepackage{amssymb}
\usepackage{amsmath}
\usepackage{amsthm}
\usepackage{booktabs}
\usepackage{enumitem}
\usepackage{subcaption}
\usepackage{amsfonts}
\usepackage{multirow}
\usepackage{ragged2e}
\usepackage{bbold}

\usepackage{newfloat}
\usepackage{listings}
\DeclareCaptionStyle{ruled}{labelfont=normalfont,labelsep=colon,strut=off} 
\floatstyle{ruled}
\newfloat{listing}{tb}{lst}{}
\floatname{listing}{Listing}
\title{FaNe: Towards Fine-Grained Cross-Modal Contrast with False-Negative Reduction and Text-Conditioned Sparse Attention}
\author{
    Peng Zhang\textsuperscript{\rm 1},
    Zhihui Lai\textsuperscript{\rm 1}\thanks{Corresponding authors},
    Wenting Chen\textsuperscript{\rm 2}\footnotemark[1],
    Xu Wu\textsuperscript{\rm 1},
    Heng Kong\textsuperscript{\rm 3}
}
\affiliations{
    \textsuperscript{\rm 1}College of Computer Science and Software Engineering, Shenzhen University\\
    \textsuperscript{\rm 2}Department of Radiation Oncology, Stanford University\\
    \textsuperscript{\rm 3}Department of Thyroid and Breast Surgery, Affiliated Hospital Group of Guangdong Medical University, Shenzhen Baoan Central Hospital

    zhang\_cloud1@163.com, laizhihui@szu.edu.cn, wentchen@stanford.edu, csxunwu@gmail.com, generaldoc@126.com
}

\usepackage{bibentry}

\begin{document}

\maketitle

\begin{abstract}
Medical vision-language pre-training (VLP) offers significant potential for advancing medical image understanding by leveraging paired image-report data. However, existing methods are limited by \textbf{Fa}lse \textbf{Ne}gatives (FaNe) induced by semantically similar texts and insufficient fine-grained cross-modal alignment. To address these limitations, we propose FaNe, a semantic-enhanced VLP framework. To mitigate false negatives, we introduce a semantic-aware positive pair mining strategy based on text-text similarity with adaptive normalization. Furthermore, we design a text-conditioned sparse attention pooling module to enable fine-grained image-text alignment through localized visual representations guided by textual cues. To strengthen intra-modal discrimination, we develop a hard-negative aware contrastive loss that adaptively reweights semantically similar negatives. Extensive experiments on five downstream medical imaging benchmarks demonstrate that FaNe achieves state-of-the-art performance across image classification, object detection, and semantic segmentation, validating the effectiveness of our framework.
\end{abstract}

\begin{links}
    \link{Code}{https://github.com/Aventador8/FaNe}
\end{links}

\section{Introduction}

Representation learning for medical radiographs has gained significant attention due to the growing availability of annotated datasets \cite{bg:mimic,bg:1}. Supervised methods excel in downstream tasks using large-scale labeled data \cite{bg:2,bg:3}, but acquiring such annotations is labor-intensive and costly, especially in the medical domain requiring expert knowledge. To overcome this, unsupervised pre-training approaches \cite{bg:4,bg:5} leverage free-text medical reports as auxiliary supervision, aligning images with text to capture high-level semantics without manual labels, offering a compelling alternative to supervised methods.

\begin{figure}
    \centering
    \includegraphics[width=\columnwidth]{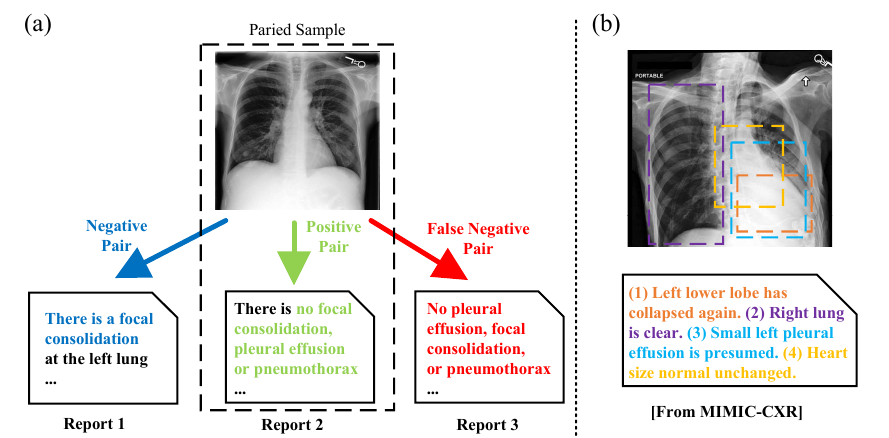}
    \caption{Illustration of false-negative problem and fine-grained alignment. (a) CLIP-style methods simply treat unpaired text or image as negatives, easily resulting in false-negative pairs (i.e, semantically similar but from different reports). (b) CLIP-style methods capture only global alignment, while color-coded sentences reveal the need for fine-grained region-level alignment. }
    \label{fig:intro}
\end{figure}

Although existing vision-language pre-training (VLP) methods have advanced performance on downstream medical tasks, they often overlook the false negative problem. In standard VLP training, an image paired with its corresponding report is treated as a positive sample, and all other reports are considered negative samples, as illustrated in Figure~\ref{fig:intro}(a). In clinical practice, different patients may exhibit the same diseases or lesions, resulting in reports with highly similar or even identical descriptions. As shown in Figure~\ref{fig:intro}(a), report 2 and report 3 contain the same findings, yet report 3 is still treated as a negative sample for the image, introducing erroneous negative pairs. In fact, the image should align with the first part of report 3, which shares similar findings with report 2, while remaining disaligned with the other part of report 3. However, current VLP methods have not considered such false negative problem, resulting in incorrect alignment during pre-training. Thus, it is essential to develop alignment strategies that can effectively mitigate the false negative issue.

\begin{figure*}[tbh]
    \centering
    \includegraphics[width=\linewidth]{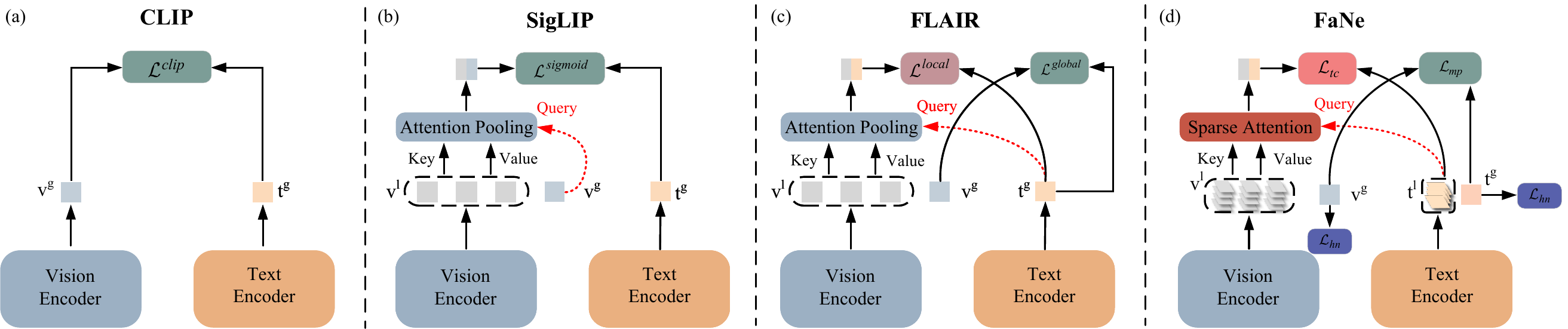}
    \caption{Comparison of text-conditioned sparse attention pooling with existing methods. (a) CLIP aligns global image $\text{v}^{\text{g}}$ and text $\text{t}^{\text{g}}$ tokens. (b) SigLIP uses a learnable global token $\text{v}^{\text{g}}$ as query to pool local tokens $\text{v}^{\text{l}}$ via cross-attention. (c) FLAIR adopts text-conditioned pooling with $\text{t}^{\text{g}}$ as query to aggregate $\text{v}^{\text{l}}$ for local alignment. (d) FaNe applies sparse attention for local text-aware features and employs a hard-negative aware contrastive loss $\mathcal{L}_{\text{hn}}$ to separate hard negatives for fine-grained understanding.}
    \label{fig:comparsion}
\end{figure*}

Another key challenge in VLP is the limited capture of detailed visual features, as CLIP performs global image-text alignment. Current VLP approaches for fine-grained alignment typically employ cross-attention to align local image and text features or map visual and textual tokens into a shared embedding space. However, these methods often depend on coarse alignment strategies that struggle to accurately capture semantically relevant local regions within the image. As illustrated in Figure \ref{fig:intro}(b), each sentence in a report generally corresponds to specific regions within the image, underscoring the necessity for fine-grained alignment conditioned on sentence-level semantics. For example, FLAIR \cite{rw:4} leverages text-conditioned attention pooling to improve alignment but struggles to accurately localize relevant image regions, as shown in Figure~\ref{fig:comparsion}(c). This is because cross-attention alone lacks the capacity to enforce precise spatial focus. Similarly, AdaMatch \cite{AdaMatch} attempts local alignment via implicit matching of image blocks to keywords, but lacks mechanisms to guarantee alignment accuracy. Thus, it is necessary to enhance fine-grained visual understanding in VLP models, as it underpins a broad spectrum of downstream tasks, including semantic segmentation and object detection.

To address these challenges, we introduce a semantic-enhanced VLP framework incorporating four key components: semantic class division, multi-positive global alignment, text-conditioned fine-grained alignment, and hard-negative intra-modal contrast. To mitigate false negatives, our \textbf{semantic class division} module employs semantic-aware adaptive normalization to uncover latent positives and categorize negatives into hard and easy subsets. To optimize multi-positive alignment within a batch, we propose \textbf{multi-positive global alignment}, which dynamically aligns a variable number of positive pairs. For precise local alignment, we propose \textbf{text-conditioned fine-grained alignment}, leveraging text-conditioned sparse attention pooling to generate localized image embeddings directly guided by textual cues. Furthermore, to enhance intra-modal discrimination, our \textbf{hard-negative intra-modal contrast} module adaptively reweights semantically similar negative instances, improving the model’s capacity to discern subtle intra-modal distinctions. In summary, our contributions can be described as follows: 
\begin{itemize}
    \item We propose a semantic-aware positive mining strategy to mitigate false negatives by identifying latent positives via text-text similarity and adaptive normalization.
    \item A text-conditioned sparse attention pooling module is designed for fine-grained image-text alignment, enabling localized visual grounding.
    \item A hard-negative aware contrastive loss is introduced, which adaptively reweights semantically similar negatives to strengthen intra-modal discrimination.

\end{itemize}

\section{Related Work}
\subsubsection{Vision-Language Pre-training.} VLP leverages cross-modal interactions between vision and language to learn joint representations without explicit supervision. CLIP \cite{clip} use a contrastive loss to match global image tokens with global text tokens (Figure \ref{fig:comparsion}(a)). However, CLIP only learns coarse-grained representation and ignores fine-grained alignment between modalities. Many works have been proposed to build fine-grained relationships between modalities, including token-level alignment \cite{rw:MGCA}, knowledge-guided multi-granularity alignment \cite{rw:MLIP}, and soft assignments allowing many-to-many mappings \cite{rw:2}. Along with attention pooling to form the global image embbedings (Figure \ref{fig:comparsion}(b)), SigLIP \cite{siglip} replaced the Softmax loss of vision-language pre-training with a Sigmoid-based loss. FLAIR \cite{rw:4} employs text-conditioned attention pooling to produce fine-grained image representations (Figure \ref{fig:comparsion}(c)). However, each sentence in a medical report typically corresponds to only a sparse local region within the image. To capture this correspondence, we propose a text-conditioned sparse attention pooling mechanism that facilitates fine-grained visual representation learning, complemented by hard-negative aware intra-modal discrimination (Figure \ref{fig:comparsion}(d)). 

\begin{figure*}[t]
    \centering
    \includegraphics[width=\linewidth]{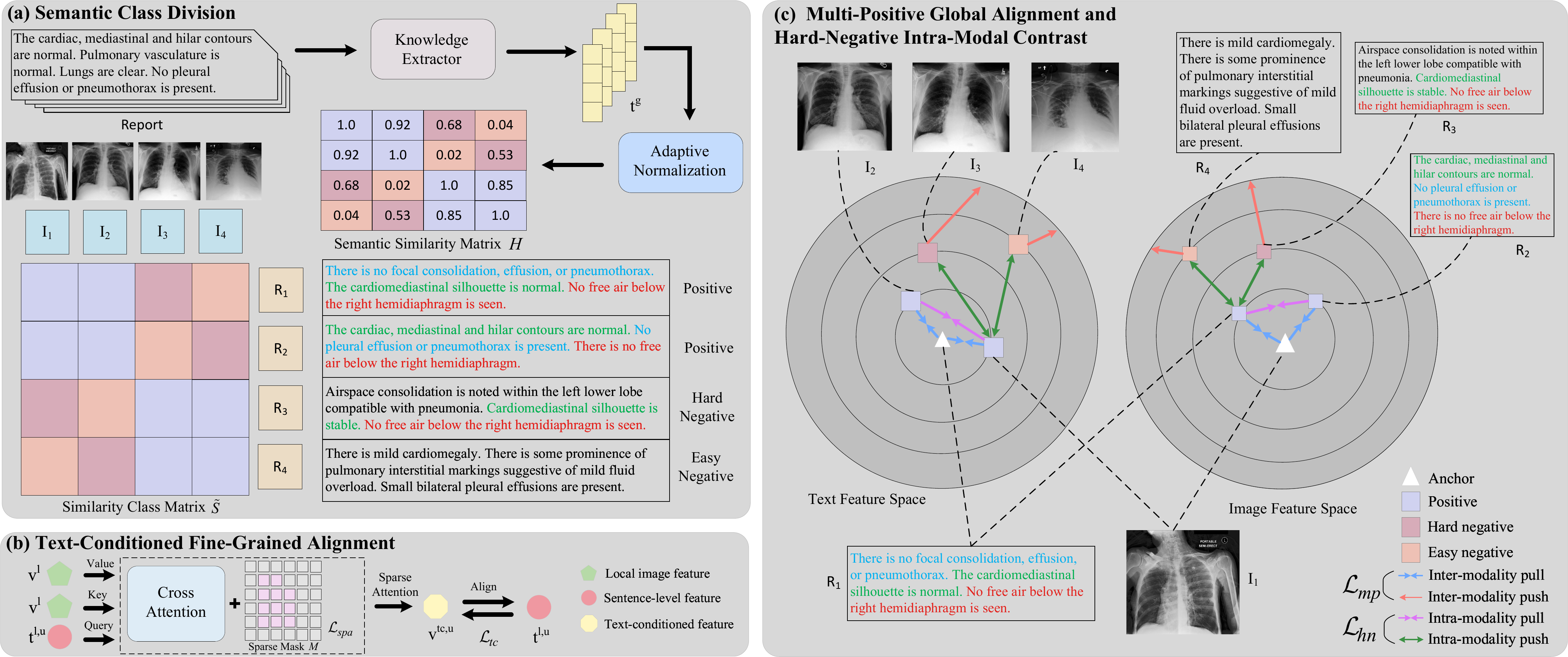}
    \caption{Overview of FaNe. (a) A semantic similarity matrix is generated via a knowledge encoder (e.g., BioClinicalBERT \cite{me:biobert}), enabling image-text pairs to be classified by a similarity threshold, with negatives further divided into hard and easy cases. (b) Text-conditioned features are derived through sparse attention, enabling alignment with sentence-level representations. (c) The combined losses $\mathcal{L}_{mp}$ and $\mathcal{L}_{hn}$ shape the feature space, where longer arrows indicate stronger repulsion for fine-grained concept learning. Sentences with the same highlight color share semantic meaning.}
    \label{fig:overview}
\end{figure*}

\subsubsection{False-Negative Problem.} Simply treating unpaired images and reports as negative pairs can lead to issues of false-negative and degrade the quality of representation. MedCLIP \cite{rw:medclip} replace the InfoNCE loss with semantic mathcing loss based on medical knowledge to eliminate false negatives. MLIP \cite{rw:MLIP} employs a knowledge-guided class-level contrastive learning strategy in conjunction with a divergence encoder to distinguish false negatives from true negatives. SAT \cite{rw:sat} computes semantic similarity between reports to group image–report pairs into positive, negative, and neutral sets, thereby avoiding the misclassification of semantically similar but unpaired samples as negatives. However, the fixed normalization of semantic similarity computation in SAT is suboptimal for varying mini-batches and different training stages. To alleviate this problem, we propose a semantic-enhanced adaptive normalization strategy to eliminate structured common semantics and stabilize training.

\section{Method}
Our overall workflow is illustrated in Figure~\ref{fig:overview}. Given reports $\text{t}$, we first use a pretrained knowledge extractor $f_{tool}$ to encode them into global text representations $\text{t}^{\text{g}}$. Next, we compute the semantic similarity matrix $\widetilde{S}$ via the semantic-enhanced adaptive normalization, followed by constructing the similarity class matrix $H$ using a predefined threshold to identify positive and negative samples. The global alignment is then optimized by applying the loss $\mathcal{L}_{mp}$. We further extract local text representations $\text{t}^{l,u}$, which are fed into a cross-attention mechanism to obtain text-conditioned visual representations $\text{v}^{\text{tc,u}}$. The fine-grained alignment loss $\mathcal{L}_{tc}$ and the sparse regularization loss $\mathcal{L}_{spa}$ are jointly applied to align $\text{t}^{l}$ with $\text{v}^{\text{tc,u}}$. Finally, for intra-modal discrimination, the hard-negative aware intra-modal contrastive loss $\mathcal{L}_{hn}$ is applied to the global image representations $\text{v}^{\text{g}}$ and global text representations $\text{t}^{\text{g}}$, enhancing the model’s ability to distinguish subtle semantic differences within each modality.

\subsection{Semantic Class Division}\label{false-negative}
Given a training set $D=\left \{ \left ( \text{v}_{1},\text{t}_{1}  \right ) ,\dots ,\left ( \text{v}_{N},\text{t}_{N}  \right ) \right \} $, where $\text{v}_{i}$ denotes a radiograph and $\text{t}_{i}$ is its corresponding report, the objective is to not only leverage ground-truth pairs $\left ( \text{v}_{i},\text{t}_{i}  \right )$ but also to correct false negatives $\left ( \text{v}_{i},\text{t}_{j}  \right )$ arising from semantically similar samples. A pretrained knowledge extractor $f_{tool}$ (e.g., BioClinicalBERT \cite{me:biobert}) encodes each report into global and local representations: $f_{tool}\left ( \text{t}_{i}  \right ) =\left [ \text{t}_{i}^{\text{l} } ,\text{t}_{i}^{\text{g} } \right ] $. To reduce bias from semantic redundancy in clinical narratives and stabilize similarity computation across batches, we introduce a semantic-enhanced adaptive normalization scheme that calibrates report-to-report similarity via batch statistics and exponential moving average (EMA) smoothing:
\begin{equation}
    p_{b}=\frac{1}{B}\sum_{i=1}^{B}\widetilde{\text{t}_{i}^{\text{g}} } 
\end{equation}
where B is the batchsize and $\widetilde{\text{t}_{i}^{g}}$ denotes unit vectors $\frac{\text{t}_{i}^{\text{g}} }{\left \|\text{t}_{i}^{\text{g}}  \right \|_{2} } $. The batch’s semantic base similarity is measured as the mean cosine similarity between each report and prototype:
\begin{equation}
    \hat{o}_{t}^{*}=\frac{1}{B}\sum_{i=1}^{B} \left \langle \widetilde{\text{t}_{i}^{\text{g}}},p_{b} \right \rangle 
\end{equation}
where $\left \langle , \right \rangle $ represents cosine similarity function. To improve training stability and prevent sudden shifts across batches, we smooth the estimate using EMA:
\begin{equation}
    o_{t}^{*} = \alpha \cdot \hat{o}_{t}^{*}+\left ( 1-\alpha  \right ) \cdot \hat{o}_{t-1}^{*}
\end{equation}
where $\alpha$ is set to 0.05, and $o_{0}^{*}=\hat{o}_{0}^{*} $. Given the normalized pairwise cosine similarity matrix $S=\left \langle \widetilde{\text{t}_{i}^{\text{g}}},\widetilde{\text{t}_{j}^{\text{g}}} \right \rangle $, we perform center-shift normalization to obtain semantic similarity matrix $\widetilde{S}$:
\begin{equation}
    \widetilde{S}=\frac{S-o_{t}^{*}}{1-o_{t}^{*}+\epsilon }
\end{equation}
where $\epsilon$ is a small constant to avoid division by zero.

After obtaining $\widetilde{S}$, we define the similarity class matrix $H\in \left \{ 0,1 \right \}^{B\times B}$ as follows:
\begin{equation}
   H=\mathbb{1}\left [ \widetilde{S}> \kappa  \right ] 
\end{equation}
where $\kappa$ is a predefined threshold. Note that $H$ re-assigns a variable number of positives to each sample.

\subsection{Multi-Positive Global Alignment}\label{multi-positive}
The symmetrical contrastive loss employed in CLIP, which facilitates bidirectional text-to-image and image-to-text alignment, is designed to handle only a single positive pair per sample. This design choice conflicts with the requirements outlined in Section \ref{false-negative}, where the proposed methods necessitate training with a variable number of positive pairs per sample. To address this limitation, we adopt a sigmoid-based contrastive loss following SigLIP \cite{siglip}, which naturally accommodates multiple positive pairs within a batch. Compared to the InfoNCE loss \cite{me:infonce}, this formulation offers greater flexibility for handling many-to-many alignments. Hence, multi-positive global alignment loss is defined:
\begin{equation}
    \mathcal{L}_{mp}= -\frac{1}{N}\sum_{i=1}^{N}\sum_{j=1}^{N}log\frac{1}{1+e^{h_{ij}\left ( -\left \langle \text{v}_{i}^{\text{g}},\text{t}_{j}^{\text{g} } \right \rangle /\tau_{1} +b  \right ) }}    
\end{equation}
where $h_{ij}$ denote the $\left ( i,j \right )$-th entry of pairwise label matrix $H$ (-1 for negative and 1 for positive pairs). To modulate similarity scaling, a temperature $\tau_{1}$ is introduced. Additionally, a learnable bias term $b$, initialized to a negative value, is incorporated to offset the imbalance between positive and negative pairs, ensuring a lower initial loss by reducing the contribution of the overwhelming number of negatives during early training.

\subsection{Text-Conditioned Fine-Grained Alignment}\label{text-conditioned}
A key distinction between medical and natural images lies in the critical role of multi-scale visual features. Medical diagnosis relies on both global anatomical context and fine-grained local cues, demanding effective integration of hierarchical representations. Moreover, medical reports often comprise multiple sentences, each describing region-specific findings (as shown in Figure \ref{fig:intro}(b)), further underscoring the need for localized visual-textual alignment. Therefore, we propose to contextualize the image representations with the individual setences, producing a unique image representation for every image-text pair.
\\
\textbf{Image and Text Representation.} Two independent encoders $f_{img}$ and $f_{txt}$ are used to extract features:
\begin{equation}
     f_{img}\left ( \text{v}  \right ) =\left [ \text{v}^{\text{l} },\text{v}^{\text{g} } \right ],f_{txt}\left ( \text{t}  \right ) =\left [ \text{t}^{\text{l} },\text{t}^{\text{g} } \right ]
\end{equation}
where $\text{v}^{\text{l}}\in\mathbb{R}^{I\times D}$ and $\text{t}^{\text{l}}\in\mathbb{R}^{P\times L\times D}$ represent local image embeddings and text embeddings, respectively. $\text{v}^{\text{g}}\in\mathbb{R}^{D}$ and $\text{t}^{\text{l}}\in\mathbb{R}^{D}$ denote global image embeddings and text embeddings, respectively. $I$ is the number of local image patches. $P$ is the total number of sentences in the report. $L$ is the maximum text length and $D$ refers to the dimension of features. In our experiments, the dimension of final output from both images and texts are the same.
\\
\textbf{Sparse Attention Mask.} Each sentence in a medical report typically corresponds to only a local region of the image. Hence, how to effectively model fine-grained semantic alignment is a unique challenge. To address it, a learnable sparse attention mask is devised:
\begin{equation}
    M=\sigma \left ( MLP\left ( [\mathcal{R\left ( \text{v}^{\text{l} }  \right ),\mathcal{R}\left (  \text{t}^{\text{l},\text{u} } \right )   } ] \right )  \right ) 
\end{equation}
where $u$ is the sentence index of each report and $M\in \mathbb{R}^{L\times I}$. $MLP\left ( \cdot  \right )$ denotes two layer linear transformations. $\mathcal{R}$ represents reshape operation. $\sigma$ is the sigmoid activation function.

To promote sparsity and reduce redundancy, we further impose the sparsity constraint overlayed on top of $M$:
\begin{equation}
    \mathcal{L}_{spa}=\sum_{i=1}^{L}\sum_{j=1}^{I}\left | M_{i,j} \right | 
\end{equation}
where $M_{i,j}$ are the activation values of the learnable attention mask $M$.
\\
\textbf{Cross-Attention Fine-Grained Alignment.} To infuse semantic context into visual representations, we employ a text-conditioned cross-attention sparse pooling mechanism. Specifically, the local sentence embedding serves as a query to aggregate local image patch embeddings, yielding a text-conditioned visual representation $\text{v}^{\text{tc,u}}$ as formulated:
\begin{equation}
    \text{v}^{\text{tc,u}}=\sum_{k=1}^{I} LN (  ( \sigma  ( \frac{\text{t}^{\text{l,u} }W_{q} ( \text{v}^{\text{l,}k}W_{k}   )^{T}  }{\sqrt{D} } \cdot M  ) \text{v}^{\text{l,}k} W_{v}  ) W_{o}  )  
    \label{cross-attention}
\end{equation}
where $W_{q},W_{k},W_{v},W_{O}$ are learnable matrices. $k$ is the spatial location index of each patch. $LN$ stands for LayerNorm.

After obtaining the $\text{v}^{\text{tc,u}}$, the InfoNCE loss is applied on $\left ( \text{v}^{\text{tc,u} },\text{t}^{\text{l,u}} \right ) $ pairs where $\text{v}^{\text{tc,u} }$ and $\text{t}^{\text{l,u}}$ from the same sentence are regarded as a positive pair, while those from different sentences are negative pairs. Given that semantically similar sentences frequently occur across different reports, we restrict the selection of negative samples for local alignment to sentences within the same report, thereby mitigating false negatives and enhancing alignment precision. We symmetrize the optimization by jointly minimizing the local text-to-image and image-to-text contrastive losses, with the total loss defined as the sum of both directional components.
\begin{equation}
    \mathcal{L}_{t2i}=- \frac{1}{\sum_{i=1}^{N}P_{i}} \sum_{i=1}^{N}\sum_{u=1}^{P_{i}}log\frac{\text{exp} ( \text{t}^{\text{l,u}}_{i} \cdot (\text{v}_{i}^{\text{tc,u}} )^{T} /\tau _{2} ) }{\sum_{k=1}^{P_{i}}\text{exp}(\text{t}^{\text{l,u}}_{i}\cdot (\text{v}_{i}^{\text{tc,k}} )^{T}/\tau_{2}) }   
\end{equation}
\begin{equation}
    \mathcal{L}_{i2t}=- \frac{1}{\sum_{i=1}^{N}P_{i}} \sum_{i=1}^{N}\sum_{u=1}^{P_{i}}log\frac{\text{exp} ( \text{t}^{\text{l,u}}_{i} \cdot (\text{v}_{i}^{\text{tc,u}} )^{T} /\tau _{2} ) }{\sum_{k=1}^{P_{i}}\text{exp}(\text{t}^{\text{l,k}}_{i}\cdot (\text{v}_{i}^{\text{tc,u}} )^{T}/\tau_{2}) }   
\end{equation}
\begin{equation}
    \mathcal{L}_{tc} =\left ( \mathcal{L}_{t2i}+\mathcal{L}_{i2t}  \right )/2
\end{equation}
where $\tau_{2}$ is the temperature parameter and $P_{i}$ denotes the number of sentences of $i$-th report.

\subsection{Hard-Negative Intra-Modal Contrast}\label{hard-negative}
The core motivation for introducing the hard-negative intra-modal contrastive loss lies in its ability to explicitly strengthen the model’s discriminative capacity for subtle semantic variations. By assigning greater importance to semantically similar negative samples, which are often underemphasized in standard contrastive learning, this loss formulation encourages the model to focus on fine-grained intra-modal distinctions. Such targeted supervision is particularly crucial in medical vision language pretraining, where accurately capturing nuanced clinical differences can significantly impact downstream diagnostic performance. So the symmetric hard-negative intra-modal contrastive loss is defined as follow:
\begin{equation}
    \mathcal{L}_{img}=-\frac{1}{N}\sum_{i=1}^{N}log\frac{1}{\sum_{j=1}^{N}\text{exp}(y_{ij}\cdot \alpha_{ij} \cdot (\text{v}^{\text{g}}_{i}\cdot (\text{v}^{\text{g}}_{j})^{T} /\tau_{3})) } 
\end{equation}
\begin{equation}
    \mathcal{L}_{text}=-\frac{1}{N}\sum_{i=1}^{N}log\frac{1}{\sum_{j=1}^{N}\text{exp}(y_{ij}\cdot \beta_{ij}  \cdot (\text{t}^{\text{g}}_{i}\cdot (\text{t}^{\text{g}}_{j})^{T} /\tau_{3})) } 
\end{equation}
\begin{equation}
   y_{ij}=\begin{cases}
 1 & \text{ if } h_{ij}=-1 \\
 0 & \text{ if } h_{ij}=
1\end{cases} 
\end{equation}
\begin{equation}
\alpha_{ij}= \frac{y_{ij}\!\cdot\! \text{v}^{\text{g}}_{i}\!\cdot\! (\text{v}^{\text{g}}_{j})^{T}/\tau_{3}}{ {\textstyle \sum_{k\neq i}^{}y_{ik}\!\cdot\!\text{v}^{\text{g}}_{i}\!\cdot\! (\text{v}^{\text{g}}_{k})^{T}/\tau_{3} } }, \beta _{ij}= \frac{y_{ij}\!\cdot\!\text{t}^{\text{g}}_{i}\!\cdot\! (\text{t}^{\text{g}}_{j})^{T}/\tau_{3}}{ {\textstyle \sum_{k\neq i}^{}y_{ik}\!\cdot\!\text{t}^{\text{g}}_{i}\!\cdot\! (\text{t}^{\text{g}}_{k})^{T}/\tau_{3} } } 
\end{equation}
\begin{equation}
    \mathcal{L}_{hn}=(\mathcal{L}_{img}+\mathcal{L}_{text})/2
\end{equation}
where $y_{ij}$ stands for the index of negative pairs and $\tau_{3}$ is the temperature parameter. Weights $\alpha_{ij}$ and $\beta_{ij}$ are designed that hard-negative pairs (i.e., higher semantic similarity in negative pairs) are underscored. This enables the model to effectively capture fine-grained semantic distinctions between true positive pairs and challenging hard-negative counterparts, thereby fostering more discriminative and concept-aware representations.

\subsection{Overall Objective}\label{overall}
The overall loss function of the proposed FaNe contains the following terms: multi-positive global alignment loss $\mathcal{L}_{mp}$, sparse regularization loss $\mathcal{L}_{spa}$, text-conditioned fine-grained alignment loss $\mathcal{L}_{tc}$ and hard-negative aware intra-modal contrastive loss $\mathcal{L}_{hn}$. We optimize these terms jointly via a weighted sum:
\begin{equation}
\mathcal{L}=\mathcal{L}_{mp}+\lambda_{1}\mathcal{L}_{tc}+\lambda_{2}\mathcal{L}_{hn}+\lambda_{3}\mathcal{L}_{spa}      
\end{equation}
where $\lambda_{1}$, $\lambda_{2}$, and $\lambda_{3}$ serve as hyper-parameters that modulate the relative contribution of each loss component.

\section{Experiments}
We pre-train the proposed FaNe framework on a large-scale medical image–report corpus, and comprehensively assess the quality of the learned visual representations across five benchmark datasets spanning three critical downstream medical imaging tasks. Furthermore, extensive ablation studies are conducted to systematically evaluate the contribution of each core component.
\subsection{Pre-Training Setup}
\textbf{Dataset.} We pre-train the proposed FaNe framework on the MIMIC-CXR v2 dataset \cite{mimic-jpg}, which comprises 377,110 chest radiographs and 227,827 corresponding medical reports. Following the preprocessing protocol in PRIOR \cite{prior}, we retain only frontal-view images to align with the evaluation setup used in downstream tasks, discarding all lateral-view images. Furthermore, to ensure sufficient textual content for semantic supervision, we filter out reports with fewer than four sentences. After preprocessing, we obtain a curated dataset of 182,475 high-quality image–report pairs for pre-training.
\\
\textbf{Implementation Details.} Following previous works, we use BioClinicalBERT \cite{me:biobert} as the text encoder. For image feature encoding, we apply ResNet50 \cite{resnet50} as the backbone. We resize all images into $224\times224$, and train our framework 50 epochs on 2 pieces of RTX 4090 GPUs with batch size of 98. The dimension $D$ is set 128 and the temperature hyper-parameters are $\tau_{1}=0.1$, $\tau_{2}=0.07$, $\tau_{3}=0.07$. The learning rate is initialized as 4e-4 and decays following a cosine policy. We set $\lambda_{1}=1$, $\lambda_{2}=1$, $\lambda_{3}=1$. More details can be found in the supplementary material.

\subsection{Downstream Tasks}
\textbf{Medical Semantic Segmentation.} We evaluate the segmentation performance of our model on the \textbf{RSNA} Pneumonia \cite{rsna} and \textbf{SIIM} Pneumothorax \cite{siim} datasets. Following the GLoRIA \cite{bg:gloria}, we fine-tune a U-Net \cite{unet} with a frozen ResNet-50 encoder initialized from our pre-trained model, while training the decoder on varying fractions of labeled data (1\%, 10\%, and 100\%). Segmentation quality is measured using the Dice similarity coefficient.
\\
\textbf{Medical Image Classification.} We evaluate the classification performance of our model on three standard benchmarks: \textbf{RSNA} Pneumonia \cite{rsna}, \textbf{COVIDx} \cite{covidx}, and \textbf{CheXpert} \cite{chexpert}. Following the linear evaluation protocol from MGCA \cite{rw:MGCA}, we freeze the pre-trained ViT-B/16 \cite{vit} or ResNet-50 backbone and train a linear classifier for downstream prediction. To assess data efficiency, experiments are conducted using 1\%, 10\%, and 100\% of the training data. Performance is measured by AUROC for RSNA and CheXpert, and classification accuracy for COVIDx, consistent with ConVIRT \cite{convirt}. Further details are included in the supplementary material.
\\
\textbf{Medical Object Detection.} We evaluate object detection capabilities of our pre-trained image encoder on the \textbf{RSNA} Pneumonia dataset \cite{rsna} and the \textbf{Object-CXR} dataset \cite{object-cxr}. Following the YOLOv3 \cite{yolov3} frozen backbone protocol, our pre-trained ResNet-50 encoder is fixed and integrated as the feature extractor, while only the detector head is fine-tuned. To investigate data efficiency, we conduct experiments using 1\%, 10\%, and 100\% of the training data. Detection performance is measured using mean Average Precision (mAP), averaged over IoU thresholds ranging from 0.4 to 0.75.
\begin{table*}[t]
\small
\centering
\begin{tabular}{cccccccccccccc}
\toprule
 &  & \multicolumn{6}{c}{Semantic Segmentation (Dice)} & \multicolumn{6}{c}{Object Detection (mAP)} \\ \midrule
Dataset & \multirow{2}{*}{FG} & \multicolumn{3}{c}{RSNA} & \multicolumn{3}{c}{SIIM} & \multicolumn{3}{c}{RSNA} & \multicolumn{3}{c}{Object CXR} \\
Method &  & 1\% & 10\% & 100\% & 1\% & 10\% & 100\% & 1\% & 10\% & 100\% & 1\% & 10\% & 100\% \\ \midrule
Random Init &  & 6.9 & 10.6 & 18.5 & 9.0 & 28.6 & 54.3 & 1.0 & 4.0 & 8.9 & $\sim$ & 0.5 & 4.4 \\
ImageNet Init &  & 34.8 & 39.9 & 64.0 & 10.2 & 35.5 & 63.5 & 3.6 & 8.0 & 15.7 & $\sim$ & 2.9 & 8.3 \\ \midrule
ConVIRT\cite{convirt} &  & 55.0 & 67.4 & 67.5 & 25.0 & 43.2 & 59.9 & 8.2 & 15.6 & 17.9 & $\sim$ & 8.6 & 15.9 \\
$\text{GLoRIA}^{\S}$ \cite{bg:gloria} & \checkmark & 60.3 & 68.7 & 68.3 & 37.4 & 57.1 & 64.0 & 11.6 & 16.1 & 24.8 & $\sim$ & 8.9 & 16.6 \\
MGCA \cite{rw:MGCA} & \checkmark & 63.0 & 68.3 & 69.8 & 49.7 & 59.3 & 64.2 & 12.9 & 16.8 & 24.9 & $\sim$ & 12.1 & 19.2 \\
M-FLAG \cite{mflag} &  & 64.6 & 69.7 & 70.5 & 52.5 & 61.2 & 64.8 & 13.7 & 17.5 & 25.4 & $\sim$ & 12.4 & 19.3 \\
MedKLIP* \cite{MedKLIP} &  & 66.2 & 69.4 & 71.9 & 50.2 & 60.8 & 63.9 & 8.9 & 16.3 & 24.5 & $\sim$ & 7.1 & 11.6 \\
PRIOR \cite{prior} & \checkmark & 66.4 & 68.3 & 72.7 & 51.2 & 59.7 & 66.3 & 15.6 & 18.5 & 25.2 & 2.9 & 15.2 & 19.8 \\
MLIP \cite{rw:MLIP} & \checkmark & 67.7 & 68.8 & 73.5 & 51.6 & 60.8 & 68.1 & \textbf{17.2} & 19.1 & 25.8 & 4.6 & \textbf{17.4} & 20.2 \\
IMITATE \cite{imitate} &  & \textbf{70.5} & 71.4 & 73.8 & 53.9 & 61.7 & 64.5 & 15.3 & 19.7 & 26.4 & 3.9 & 12.7 & 20.3 \\
\textbf{FaNe (ours)} & \checkmark & 69.5 & \textbf{72.4} & \textbf{74.1} & \textbf{54.1} & \textbf{62.3} & \textbf{68.8} & 16.4 & \textbf{20.6} & \textbf{27.2} & \textbf{4.9} & 15.6 & \textbf{21.2} \\ \bottomrule
\end{tabular}
\caption{Results of semantic segmentation on RSNA and SIIM datasets and object detection on RSNA and Object CXR datasets. Method marked with * uses extra annotated data and the "$\sim$" denotes mAP is samller than 1\%. The "FG" means these methods have or do not have fine-grained alignment. The $\S$ symbol denotes models pretrained on the MIMIC-CXR dataset.}
\label{SegAndDec}
\end{table*}
\begin{figure}[t]
    \centering
    \includegraphics[width=\linewidth]{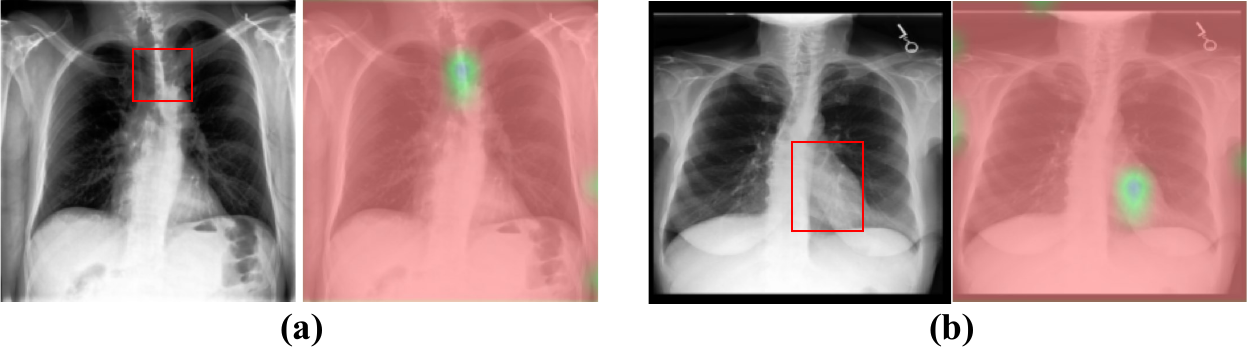}
    \caption{Representative cross-modality attention maps. (a) The related sentence is "Mild \textbf{thoracic scoliosis} noted." (b) The related sentence is "\textbf{Heart} size is normal."}
    \label{fig:attention}
\end{figure}
\subsection{Results}
The results of competing methods are cited from their respective original publications. For a fair comparison, we follow the MGCA protocol to pre-train GLoRIA on the MIMIC-CXR dataset, as indicated by the $\S$ symbol.
\\
\textbf{Results on Medical Semantic Segmentation.} We evaluate the effect of fine-grained local representations on segmentation using the ResNet-50 U-Net across the RSNA and SIIM datasets, as shown in Table~\ref{SegAndDec}. While prior methods (e.g., GLoRIA, MGCA, PRIOR, MedKLIP, MLIP) incorporate localized features during pre-training, our FaNe framework consistently outperforms them. This improvement stems from the proposed text-conditioned sparse attention pooling, which enables more precise region-level feature extraction essential for accurate segmentation.
\\
\textbf{Results on Medical Object Detection.} To further validate the effectiveness of our proposed FaNe framework, we perform object detection experiments utilizing a ResNet-50-YOLOv3 architecture on the RSNA and Object CXR datasets. As presented in Table \ref{SegAndDec}, FaNe consistently surpasses competing methods across most scenarios. We hypothesize that this superior performance stems from the strategic mining of potential positive samples through semantically aligned textual cues, which mitigates the adverse effects of misclassified negative samples. Consequently, this approach yields more precise and robust visual feature representations, enhancing the reliability of bounding box regression for object detection tasks in medical imaging.
\\
\textbf{Results on Medical Image Classification.} We perform supervised classification on the RSNA, CheXpert, and COVIDx datasets. Specifically, we pre-train our proposed FaNe model using both ResNet-50 and ViT-B/16 architectures to evaluate the impact of different image encoders. As reported in Table \ref{result:cls}, FaNe with the ViT-B/16 encoder achieves superior performance across multiple datasets, suggesting that Transformer-based architectures may be particularly effective for aligning vision and language modalities in medical imaging tasks. Notably, while existing approaches achieve promising performance, they overlook hard-negative mining, leading to suboptimal representation learning. In contrast, our proposed hard-negative aware contrastive loss imposes stronger separation on semantically similar samples, encouraging the model to learn clearer semantic boundaries. This helps reduce misclassification of boundary-ambiguous categories and enhances both robustness and recognition capability.
\begin{table*}[tb]
\small
\centering
\begin{tabular}{ccccccccccc}
\toprule
Dataset &\multirow{2}{*}{FG} & \multicolumn{3}{c}{CheXpert (AUC)} & \multicolumn{3}{c}{RSNA (AUC)} & \multicolumn{3}{c}{COVIDx (ACC)} \\
Method & & 1\% & 10\% & 100\% & 1\% & 10\% & 100\% & 1\% & 10\% & 100\% \\ \midrule
Random Init & & 56.1 & 62.6 & 65.7 & 58.9 & 69.4 & 74.1 & 50.5 & 60.3 & 70.0 \\
ImageNet Init & & 74.4 & 79.7 & 81.4 & 74.9 & 74.5 & 76.3 & 64.8 & 78.8 & 86.3 \\ \midrule
ConVIRT \cite{convirt} & & 85.9 & 86.8 & 87.3 & 77.4 & 80.1 & 81.3 & 72.5 & 82.5 & 92.0 \\
$\text{GLoRIA}^{\S}$ \cite{bg:gloria} & \checkmark& 87.1 & 88.7 & 88.0 & 87.0 & 89.4 & 90.2 & 66.5 & 80.5 & 88.8 \\
MGCA \cite{rw:MGCA} & \checkmark & 87.6 & 88.0 & 88.2 & 88.6 & 89.1 & 89.9 & 72.0 & 83.5 & 90.5 \\
PRIOR \cite{prior} & \checkmark & 87.6 & 88.6 & 88.8 & 88.9 & 89.5 & 90.5 & 72.3 & 84.7 & 91.0 \\
MedKLIP* \cite{MedKLIP} & & 86.2 & 86.5 & 87.7 & 87.3 & 88.0 & 89.3 & 74.5 & 85.2 & 90.3 \\
MLIP \cite{rw:MLIP} & \checkmark & 87.8 & 88.7 & 88.9 & 88.8 & 89.6 & 90.6 & 73.0 & 85.0 & 90.8 \\
MedUnifer \cite{medunifier} & & - & - & - & 87.6 & 88.8 & 91.7 & 76.8 & 88.3 & 93.5 \\
SENSE \cite{sense} & & 87.6 & 88.5 & 89.3 & 88.0 & 89.6 & 90.7 & - & - & - \\
FaNe (Ours, ResNet-50) & \checkmark & 88.2 & 89.1 & 89.9 & 88.9 & 89.8 & 92.6 & 78.2 & 89.1 & 94.0 \\
\textbf{FaNe (Ours, ViT-B/16)} & \checkmark & \textbf{89.7} & \textbf{90.4} & \textbf{90.8} & \textbf{89.3} & \textbf{90.2} & \textbf{93.1} & \textbf{79.5} & \textbf{90.7} & \textbf{95.5} \\ \bottomrule
\end{tabular}
\caption{Image classification performance with fine-tuning on 1\%, 10\%, and 100\% of training data for CheXpert, RSNA, and COVIDx datasets. Method marked with * uses extra annotated data. The "FG" means these methods have or do not have fine-grained alignment. The $\S$ symbol denotes models pretrained on the MIMIC-CXR dataset.}
\label{result:cls}
\end{table*}
\begin{table}[tb]
\scriptsize
\centering
\begin{tabular}{cccccccc}
\hline
\multirow{2}{*}{Learnable Mask} & \multirow{2}{*}{$\mathcal{L}_{spa}$} & \multicolumn{3}{c}{RSNA (Dice)} & \multicolumn{3}{c}{Object CXR (mAP)} \\
 &  & 1\% & 10\% & 100\% & 1\% & 10\% & 100\% \\ \hline
  &   & 66.6 & 68.9 & 71.2 & 2.2 & 12.6 & 18.5 \\
\checkmark &   & 67.7 & 70.3 & 72.6 & 3.4 & 13.8 & 19.9 \\
\checkmark & \checkmark & \textbf{69.5} & \textbf{72.4} & \textbf{74.1} & \textbf{4.9} & \textbf{15.6} & \textbf{21.2} \\ \hline
\end{tabular}
\caption{Ablation study on learnable sparse attention mask and sparse regularization loss.}
\label{sparse-attention}
\end{table}
\subsection{Ablation Study}
\textbf{Visualization.} To demonstrate the effectiveness and interpretability of our text-conditioned fine-grained alignment, we visualize cross-modality attention maps generated from image-text pairs. Using Equation \ref{cross-attention}, we compute the attention maps and resize them for overlay on the input images. As shown in Figure \ref{fig:attention}, our sparse attention pooling module with ViT as the encoder accurately highlights semantically relevant regions such as thoracic scoliosis and the heart location based on the corresponding sentence. Additional visualizations using ResNet are presented in the appendix.
\\
\textbf{Effectiveness of Sparse Attention Mask.} To assess the impact of incorporating a sparse attention mask into the cross-attention mechanism, we conduct an ablation study presented in Table~\ref{sparse-attention}. The first row reports the baseline model without any learnable attention mask. The second row introduces a learnable attention mask without applying sparsity regularization. The final row corresponds to our full method with both learnable attention and a sparsity constraint. The results clearly demonstrate that both components contribute significantly to performance gains in semantic segmentation and object detection, validating the effectiveness of our fine-grained alignment strategy.
\begin{table}[t]
\scriptsize
\centering
\begin{tabular}{ccccccc}
\hline
\multirow{2}{*}{\begin{tabular}[c]{@{}c@{}}Semantic\\ Normalization\end{tabular}} & \multicolumn{3}{c}{RSNA (Dice)} & \multicolumn{3}{c}{Object CXR (mAP)} \\
 & 1\% & 10\% & 100\% & 1\% & 10\% & 100\% \\ \hline
  & 67.1 & 69.7 & 71.5 & 2.5 & 12.9 & 18.7 \\
\checkmark & \textbf{69.5} & \textbf{72.4} & \textbf{74.1} & \textbf{4.9} & \textbf{15.6} & \textbf{21.2} \\ \hline
\end{tabular}
\caption{Ablation study on semantic-enhanced adaptive normalization.}
\label{normalization}
\end{table}
\begin{table}[t]
\centering
\tiny
\begin{tabular}{ccccccccc}
\toprule
\multicolumn{3}{c}{Components} & \multicolumn{3}{c}{RSNA (Dice)} & \multicolumn{3}{c}{SIIM (Dice)} \\
$\mathcal{L}_{mp}$ & $\mathcal{L}_{hn}$ & $\mathcal{L}_{tc}\!+\!\mathcal{L}_{spa}$ & 1\% & 10\% & 100\% & 1\% & 10\% & 100\% \\ \midrule
\checkmark &  &  & 66.7 & 69.6 & 71.0 & 50.9 & 59.3 & 64.9 \\
\checkmark & \checkmark &  & 67.8 & 70.8 & 72.3 & 52.1 & 60.5 & 66.1 \\
\checkmark &  & \checkmark & 68.2 & 71.3 & 73.1 & 52.9 & 61.2 & 67.6 \\
\checkmark & \checkmark & \checkmark & \textbf{69.5} & \textbf{72.4} & \textbf{74.1} & \textbf{54.1} & \textbf{62.3} & \textbf{68.8} \\ \bottomrule
\end{tabular}
\caption{Ablation study on losses. $\mathcal{L}_{tc}$ and $\mathcal{L}_{spa}$ are fixed combination as reported in Table \ref{sparse-attention}.}
\label{loss}
\end{table}
\begin{table}[t]
\centering
\scriptsize
\begin{tabular}{cccccc}
\hline
\multirow{2}{*}{RSNA (Dice)} & \multicolumn{5}{c}{Threshold} \\ \cmidrule{2-6} 
 & 0.6 & 0.7 & 0.8 & 0.9 & 0.95 \\ \midrule
1\% & 65.7 & 66.6 & 67.7 & 69.1 & \textbf{69.5} \\
10\% & 68.4 & 69.3 & 70.8 & 71.8 & \textbf{72.4} \\
100\% & 70.5 & 70.9 & 71.2 & 72.2 & \textbf{74.1} \\ \hline
\end{tabular}
\caption{Ablation study on semantic threshold $\kappa$.}
\label{threshold}
\end{table}
\\
\textbf{Effectiveness of Semantic-Enhanced Adaptive Normalization.} We conduct an ablation study by toggling this component on and off. As shown in Table \ref{normalization}, semantic normalization consistently improves performance across both segmentation and detection tasks under different training data ratios. These consistent gains indicate that normalization mechanism effectively mitigates the influence of semantically similar false negatives, enhancing representation learning.
\\
\textbf{Effectiveness of Losses.} As shown in Table \ref{loss}, training with the semantic-aware multi-positive global alignment loss ($\mathcal{L}_{mp}$) alone provides a strong baseline by alleviating false negatives caused by semantically similar reports. Adding the hard-negative aware contrastive loss ($\mathcal{L}_{hn}$) further improves performance, demonstrating its role in enhancing intra-modal discrimination by pushing apart semantically close but incorrect negatives. Incorporating the text-conditioned sparse contrastive loss and its sparsity regularization ($\mathcal{L}_{tc} + \mathcal{L}_{spa}$) leads to additional gains, particularly in low data ratio, validating its ability to enable fine-grained and text-guided local alignment. The combination of all components yields the highest Dice scores across both datasets, confirming the complementary benefits of our design in addressing false negatives, improving alignment precision, and refining intra-modal semantic separation.
\\
\textbf{Effectiveness of Threshold.} As shown in Table \ref{threshold}, model performance consistently improves as the semantic threshold $\kappa$ increases from 0.6 to 0.95, indicating that a higher threshold helps mitigate the impact of false-negative pairs and enhances the model’s discriminative capability. The best segmentation results are achieved when $\kappa$ is set to 0.95.

\section{Conclusion}
We present a semantic-enhanced framework to address key challenges in medical VLP, including false negatives from semantically similar reports, coarse image-text alignment, and insufficient intra-modal discrimination. Our approach integrates a semantic-aware positive pair mining strategy, a text-conditioned sparse attention pooling module, and a hard-negative aware contrastive loss to enhance the quality and granularity of learned representations. Extensive experiments on five downstream medical imaging tasks, covering classification, detection, and segmentation, demonstrate the superiority of our method over existing VLP baselines.

\section*{Acknowledgements}
This work was supported in part by the Natural Science Foundation of China under Grant 62476175 and Grant 62272319, and in part by the Natural Science Foundation of Guangdong Province (Grant 2024A1515011637, 2023B1212060076), the Science and Technology Planning Project of Shenzhen Municipality under Grants JCYJ20220818095803007 and JCYJ20250604145234045, and the Special Project for Clinical and Basic Sci\&Tech Innovation of Guangdong Medical University under Grant GDMULCJC2024151.

\bibliography{aaai2026}

\begin{thebibliography}{38}
\providecommand{\natexlab}[1]{#1}

\bibitem[{Alsentzer et~al.(2019)Alsentzer, Murphy, Boag, Weng, Jindi, Naumann, and McDermott}]{me:biobert}
Alsentzer, E.; Murphy, J.; Boag, W.; Weng, W.-H.; Jindi, D.; Naumann, T.; and McDermott, M. 2019.
\newblock Publicly Available Clinical BERT Embeddings.
\newblock In \emph{Proceedings of the 2nd Clinical Natural Language Processing Workshop}, 72--78.

\bibitem[{Chambon et~al.(2024)Chambon, Delbrouck, Sounack, Huang, Chen, Varma, Truong, Langlotz et~al.}]{bg:1}
Chambon, P.; Delbrouck, J.-B.; Sounack, T.; Huang, S.-C.; Chen, Z.; Varma, M.; Truong, S.~Q.; Langlotz, C.~P.; et~al. 2024.
\newblock Chexpert plus: Hundreds of thousands of aligned radiology texts, images and patients.
\newblock \emph{arXiv e-prints}, arXiv--2405.

\bibitem[{Chen et~al.(2024)Chen, Shen, Lin, Luo, Li, and Yuan}]{AdaMatch}
Chen, W.; Shen, L.; Lin, J.; Luo, J.; Li, X.; and Yuan, Y. 2024.
\newblock Fine-Grained Image-Text Alignment in Medical Imaging Enables Explainable Cyclic Image-Report Generation.
\newblock In \emph{Proceedings of the 62nd Annual Meeting of the Association for Computational Linguistics (Volume 1: Long Papers)}, 9494--9509.

\bibitem[{Cheng et~al.(2023)Cheng, Lin, Lyu, Huang, Luo, and Tang}]{prior}
Cheng, P.; Lin, L.; Lyu, J.; Huang, Y.; Luo, W.; and Tang, X. 2023.
\newblock Prior: Prototype representation joint learning from medical images and reports.
\newblock In \emph{Proceedings of the IEEE/CVF International Conference on Computer Vision}, 21361--21371.

\bibitem[{Chiu et~al.(2025)Chiu, Li, Chi, and Tseng}]{bg:3}
Chiu, T.-M.; Li, Y.-C.; Chi, I.-C.; and Tseng, M.-H. 2025.
\newblock AI-Driven Enhancement of Skin Cancer Diagnosis: A Two-Stage Voting Ensemble Approach Using Dermoscopic Data.
\newblock \emph{Cancers}, 17(1): 137.

\bibitem[{Dosovitskiy et~al.(2020)Dosovitskiy, Beyer, Kolesnikov, Weissenborn, Zhai, Unterthiner, Dehghani, Minderer, Heigold, Gelly et~al.}]{vit}
Dosovitskiy, A.; Beyer, L.; Kolesnikov, A.; Weissenborn, D.; Zhai, X.; Unterthiner, T.; Dehghani, M.; Minderer, M.; Heigold, G.; Gelly, S.; et~al. 2020.
\newblock An image is worth 16x16 words: Transformers for image recognition at scale.
\newblock \emph{arXiv preprint arXiv:2010.11929}.

\bibitem[{Gao et~al.(2024)Gao, Liu, Xu, Wu, Zhang, Li, Yang, Liu, and Sun}]{rw:2}
Gao, Y.; Liu, J.; Xu, Z.; Wu, T.; Zhang, E.; Li, K.; Yang, J.; Liu, W.; and Sun, X. 2024.
\newblock Softclip: Softer cross-modal alignment makes clip stronger.
\newblock In \emph{Proceedings of the AAAI Conference on Artificial Intelligence}, volume~38, 1860--1868.

\bibitem[{He et~al.(2016)He, Zhang, Ren, and Sun}]{resnet50}
He, K.; Zhang, X.; Ren, S.; and Sun, J. 2016.
\newblock Deep residual learning for image recognition.
\newblock In \emph{Proceedings of the IEEE conference on computer vision and pattern recognition}, 770--778.

\bibitem[{Healthcare(2020)}]{object-cxr}
Healthcare, J. 2020.
\newblock Object-cxr-automatic detection of foreign objects on chest x-rays.
\newblock \emph{MIDL}.

\bibitem[{Huang et~al.(2021)Huang, Shen, Lungren, and Yeung}]{bg:gloria}
Huang, S.-C.; Shen, L.; Lungren, M.~P.; and Yeung, S. 2021.
\newblock Gloria: A multimodal global-local representation learning framework for label-efficient medical image recognition.
\newblock In \emph{Proceedings of the IEEE/CVF international conference on computer vision}, 3942--3951.

\bibitem[{Irvin et~al.(2019)Irvin, Rajpurkar, Ko, Yu, Ciurea-Ilcus, Chute, Marklund, Haghgoo, Ball, Shpanskaya et~al.}]{chexpert}
Irvin, J.; Rajpurkar, P.; Ko, M.; Yu, Y.; Ciurea-Ilcus, S.; Chute, C.; Marklund, H.; Haghgoo, B.; Ball, R.; Shpanskaya, K.; et~al. 2019.
\newblock Chexpert: A large chest radiograph dataset with uncertainty labels and expert comparison.
\newblock In \emph{Proceedings of the AAAI conference on artificial intelligence}, volume~33, 590--597.

\bibitem[{Johnson et~al.(2019{\natexlab{a}})Johnson, Pollard, Berkowitz, Greenbaum, Lungren, Deng, Mark, and Horng}]{bg:mimic}
Johnson, A.~E.; Pollard, T.~J.; Berkowitz, S.~J.; Greenbaum, N.~R.; Lungren, M.~P.; Deng, C.-y.; Mark, R.~G.; and Horng, S. 2019{\natexlab{a}}.
\newblock MIMIC-CXR, a de-identified publicly available database of chest radiographs with free-text reports.
\newblock \emph{Scientific data}, 6(1): 317.

\bibitem[{Johnson et~al.(2019{\natexlab{b}})Johnson, Pollard, Greenbaum, Lungren, Deng, Peng, Lu, Mark, Berkowitz, and Horng}]{mimic-jpg}
Johnson, A.~E.; Pollard, T.~J.; Greenbaum, N.~R.; Lungren, M.~P.; Deng, C.-y.; Peng, Y.; Lu, Z.; Mark, R.~G.; Berkowitz, S.~J.; and Horng, S. 2019{\natexlab{b}}.
\newblock MIMIC-CXR-JPG, a large publicly available database of labeled chest radiographs.
\newblock \emph{arXiv preprint arXiv:1901.07042}.

\bibitem[{Lai et~al.(2024)Lai, Yao, Jiang, Wang, He, Tao, and Zhou}]{bg:5}
Lai, H.; Yao, Q.; Jiang, Z.; Wang, R.; He, Z.; Tao, X.; and Zhou, S.~K. 2024.
\newblock Carzero: Cross-attention alignment for radiology zero-shot classification.
\newblock In \emph{Proceedings of the IEEE/CVF Conference on Computer Vision and Pattern Recognition}, 11137--11146.

\bibitem[{Li et~al.(2024)Li, Yang, Ren, Nie, Gao, Tan, and Li}]{rw:MLIP}
Li, Z.; Yang, L.~T.; Ren, B.; Nie, X.; Gao, Z.; Tan, C.; and Li, S.~Z. 2024.
\newblock Mlip: Enhancing medical visual representation with divergence encoder and knowledge-guided contrastive learning.
\newblock In \emph{Proceedings of the IEEE/CVF Conference on Computer Vision and Pattern Recognition}, 11704--11714.

\bibitem[{Liu et~al.(2023{\natexlab{a}})Liu, Lu, Wei, Wu, Wang, Zhang, and Zheng}]{rw:sat}
Liu, B.; Lu, D.; Wei, D.; Wu, X.; Wang, Y.; Zhang, Y.; and Zheng, Y. 2023{\natexlab{a}}.
\newblock Improving medical vision-language contrastive pretraining with semantics-aware triage.
\newblock \emph{IEEE Transactions on Medical Imaging}, 42(12): 3579--3589.

\bibitem[{Liu, Lu, and Wang(2024)}]{sense}
Liu, B.; Lu, Z.; and Wang, Y. 2024.
\newblock Towards Medical Vision-Language Contrastive Pre-training via Study-Oriented Semantic Exploration.
\newblock In \emph{Proceedings of the 32nd ACM International Conference on Multimedia}, 4861--4870.

\bibitem[{Liu et~al.(2023{\natexlab{b}})Liu, Cheng, Chen, Qiao, Zhang, Shah, Bai, and Arcucci}]{mflag}
Liu, C.; Cheng, S.; Chen, C.; Qiao, M.; Zhang, W.; Shah, A.; Bai, W.; and Arcucci, R. 2023{\natexlab{b}}.
\newblock M-flag: Medical vision-language pre-training with frozen language models and latent space geometry optimization.
\newblock In \emph{International Conference on Medical Image Computing and Computer-Assisted Intervention}, 637--647. Springer.

\bibitem[{Liu et~al.(2024)Liu, Cheng, Shi, Shah, Bai, and Arcucci}]{imitate}
Liu, C.; Cheng, S.; Shi, M.; Shah, A.; Bai, W.; and Arcucci, R. 2024.
\newblock Imitate: Clinical prior guided hierarchical vision-language pre-training.
\newblock \emph{IEEE Transactions on Medical Imaging}.

\bibitem[{Marcel and Rodriguez(2010)}]{torchvision}
Marcel, S.; and Rodriguez, Y. 2010.
\newblock Torchvision the machine-vision package of torch.
\newblock In \emph{Proceedings of the 18th ACM international conference on Multimedia}, 1485--1488.

\bibitem[{Oord, Li, and Vinyals(2018)}]{me:infonce}
Oord, A. v.~d.; Li, Y.; and Vinyals, O. 2018.
\newblock Representation learning with contrastive predictive coding.
\newblock \emph{arXiv preprint arXiv:1807.03748}.

\bibitem[{Radford et~al.(2021)Radford, Kim, Hallacy, Ramesh, Goh, Agarwal, Sastry, Askell, Mishkin, Clark et~al.}]{clip}
Radford, A.; Kim, J.~W.; Hallacy, C.; Ramesh, A.; Goh, G.; Agarwal, S.; Sastry, G.; Askell, A.; Mishkin, P.; Clark, J.; et~al. 2021.
\newblock Learning transferable visual models from natural language supervision.
\newblock In \emph{International conference on machine learning}, 8748--8763. PmLR.

\bibitem[{Redmon and Farhadi(2018)}]{yolov3}
Redmon, J.; and Farhadi, A. 2018.
\newblock Yolov3: An incremental improvement.
\newblock \emph{arXiv preprint arXiv:1804.02767}.

\bibitem[{Ronneberger, Fischer, and Brox(2015)}]{unet}
Ronneberger, O.; Fischer, P.; and Brox, T. 2015.
\newblock U-net: Convolutional networks for biomedical image segmentation.
\newblock In \emph{Medical image computing and computer-assisted intervention--MICCAI 2015: 18th international conference, Munich, Germany, October 5-9, 2015, proceedings, part III 18}, 234--241. Springer.

\bibitem[{Russakovsky et~al.(2015)Russakovsky, Deng, Su, Krause, Satheesh, Ma, Huang, Karpathy, Khosla, Bernstein et~al.}]{imagenet}
Russakovsky, O.; Deng, J.; Su, H.; Krause, J.; Satheesh, S.; Ma, S.; Huang, Z.; Karpathy, A.; Khosla, A.; Bernstein, M.; et~al. 2015.
\newblock Imagenet large scale visual recognition challenge.
\newblock \emph{International journal of computer vision}, 115: 211--252.

\bibitem[{Shih et~al.(2019)Shih, Wu, Halabi, Kohli, Prevedello, Cook, Sharma, Amorosa, Arteaga, Galperin-Aizenberg et~al.}]{rsna}
Shih, G.; Wu, C.~C.; Halabi, S.~S.; Kohli, M.~D.; Prevedello, L.~M.; Cook, T.~S.; Sharma, A.; Amorosa, J.~K.; Arteaga, V.; Galperin-Aizenberg, M.; et~al. 2019.
\newblock Augmenting the national institutes of health chest radiograph dataset with expert annotations of possible pneumonia.
\newblock \emph{Radiology: Artificial Intelligence}, 1(1): e180041.

\bibitem[{Wang et~al.(2022{\natexlab{a}})Wang, Zhou, Wang, Vardhanabhuti, and Yu}]{rw:MGCA}
Wang, F.; Zhou, Y.; Wang, S.; Vardhanabhuti, V.; and Yu, L. 2022{\natexlab{a}}.
\newblock Multi-granularity cross-modal alignment for generalized medical visual representation learning.
\newblock In \emph{Proceedings of International Conference on Neural Information Processing Systems}, 33536--33549.

\bibitem[{Wang, Lin, and Wong(2020)}]{covidx}
Wang, L.; Lin, Z.~Q.; and Wong, A. 2020.
\newblock Covid-net: A tailored deep convolutional neural network design for detection of covid-19 cases from chest x-ray images.
\newblock \emph{Scientific reports}, 10(1): 19549.

\bibitem[{Wang et~al.(2022{\natexlab{b}})Wang, Wu, Agarwal, and Sun}]{rw:medclip}
Wang, Z.; Wu, Z.; Agarwal, D.; and Sun, J. 2022{\natexlab{b}}.
\newblock Medclip: Contrastive learning from unpaired medical images and text.
\newblock In \emph{Proceedings of the Conference on Empirical Methods in Natural Language Processing. Conference on Empirical Methods in Natural Language Processing}, volume 2022, 3876.

\bibitem[{Wolf et~al.(2020)Wolf, Debut, Sanh, Chaumond, Delangue, Moi, Cistac, Rault, Louf, Funtowicz et~al.}]{transformer-library}
Wolf, T.; Debut, L.; Sanh, V.; Chaumond, J.; Delangue, C.; Moi, A.; Cistac, P.; Rault, T.; Louf, R.; Funtowicz, M.; et~al. 2020.
\newblock Transformers: State-of-the-art natural language processing.
\newblock In \emph{Proceedings of the 2020 conference on empirical methods in natural language processing: system demonstrations}, 38--45.

\bibitem[{Wu et~al.(2023)Wu, Zhang, Zhang, Wang, and Xie}]{MedKLIP}
Wu, C.; Zhang, X.; Zhang, Y.; Wang, Y.; and Xie, W. 2023.
\newblock Medklip: Medical knowledge enhanced language-image pre-training for x-ray diagnosis.
\newblock In \emph{Proceedings of the IEEE/CVF International Conference on Computer Vision}, 21372--21383.

\bibitem[{Wu et~al.(2024)Wu, Lai, Zhou, Hou, Pedrycz, and Shen}]{bg:2}
Wu, X.; Lai, Z.; Zhou, J.; Hou, X.; Pedrycz, W.; and Shen, L. 2024.
\newblock Light-aware contrastive learning for low-light image enhancement.
\newblock \emph{ACM Transactions on Multimedia Computing, Communications and Applications}, 20(9): 1--24.

\bibitem[{Xiao et~al.(2025)Xiao, Kim, Georgescu, Akata, and Alaniz}]{rw:4}
Xiao, R.; Kim, S.; Georgescu, M.-I.; Akata, Z.; and Alaniz, S. 2025.
\newblock FLAIR: VLM with Fine-grained Language-informed Image Representations.
\newblock In \emph{Proceedings of the IEEE/CVF Conference on Computer Vision and Pattern Recognition}.

\bibitem[{Zawacki et~al.(2019)Zawacki, Wu, Shih, Elliott, Fomitchev, Hussain, ParasLakhani, Culliton, and Bao}]{siim}
Zawacki, A.; Wu, C.; Shih, G.; Elliott, J.; Fomitchev, M.; Hussain, M.; ParasLakhani; Culliton, P.; and Bao, S. 2019.
\newblock SIIM-ACR Pneumothorax Segmentation.
\newblock \url{https://kaggle.com/competitions/siim-acr-pneumothorax-segmentation}.
\newblock Kaggle.

\bibitem[{Zhai et~al.(2023)Zhai, Mustafa, Kolesnikov, and Beyer}]{siglip}
Zhai, X.; Mustafa, B.; Kolesnikov, A.; and Beyer, L. 2023.
\newblock Sigmoid loss for language image pre-training.
\newblock In \emph{Proceedings of the IEEE/CVF international conference on computer vision}, 11975--11986.

\bibitem[{Zhang et~al.(2023)Zhang, Wu, Zhang, Xie, and Wang}]{bg:4}
Zhang, X.; Wu, C.; Zhang, Y.; Xie, W.; and Wang, Y. 2023.
\newblock Knowledge-enhanced visual-language pre-training on chest radiology images.
\newblock \emph{Nature Communications}, 14(1): 4542.

\bibitem[{Zhang et~al.(2022)Zhang, Jiang, Miura, Manning, and Langlotz}]{convirt}
Zhang, Y.; Jiang, H.; Miura, Y.; Manning, C.~D.; and Langlotz, C.~P. 2022.
\newblock Contrastive learning of medical visual representations from paired images and text.
\newblock In \emph{Machine learning for healthcare conference}, 2--25. PMLR.

\bibitem[{Zhang et~al.(2025)Zhang, Yu, Chen, Yang, and Yeo}]{medunifier}
Zhang, Z.; Yu, Y.; Chen, Y.; Yang, X.; and Yeo, S.~Y. 2025.
\newblock Medunifier: Unifying vision-and-language pre-training on medical data with vision generation task using discrete visual representations.
\newblock \emph{arXiv preprint arXiv:2503.01019}.

\end{thebibliography}
\newpage
\twocolumn[
\begin{center}
    \Huge\bfseries Supplementary Material \par
\end{center}
]

\appendix
\section{Appendix}
\subsection{Data Preprocessing}
We pre-train the proposed FaNe framework on the MIMIC-CXR-JPG dataset. To ensure consistency and quality, lateral-view radiographs and image-report pairs with fewer than four sentences in the associated report are excluded. After filtering, the resulting dataset comprises 182,475 radiograph-report pairs.\\
\\
\textbf{Report Preprocessing.} We extract all sentences from the \textit{Findings} and \textit{Impression} sections of each report. The texts are tokenized using the tokenizer from BioClinicalBERT \cite{me:biobert}, implemented via the Hugging Face Transformers library \cite{transformer-library}. Each tokenized report is padded to a fixed length of 112 tokens to enable batch processing.\\
\\
\textbf{Image Preprocessing.} All radiographs are normalized to the range [0,1] using a mean of [0.4755, 0.4755, 0.406] and a standard deviation of [0.229, 0.224, 0.225]. Subsequently, each image is resized to a resolution of $224 \times 224$ pixels before being input into the model.

\subsection{Image and Text Encoder} 
\textbf{Image Encoder.} By default, we adopt ResNet-50 as the image encoder, utilizing the implementation provided by the torchvision library \cite{torchvision}, with weights initialized from ImageNet \cite{imagenet}. Following the preprocessing strategy in Gloria \cite{bg:gloria}, each radiograph is upsampled from $224 \times 224$ to $299 \times 299$ before being input to the network. For global representation, we apply an attention-based pooling layer over the final feature maps to derive a 2048-dimensional image-level embedding. For local representation, we extract intermediate features from the third bottleneck block of ResNet-50, yielding a feature tensor $f \in \mathbb{R}^{1024 \times 19 \times 19}$. This tensor is subsequently reshaped into $f \in \mathbb{R}^{1024 \times 361}$, where each of the 361 vectors corresponds to a distinct image region.

To assess the influence of backbone architecture, we additionally experiment with the ViT-B/16 vision transformer \cite{vit}, initialized with ImageNet-1k pre-trained weights. In accordance with standard ViT protocols, each input image is partitioned into non-overlapping patches of size $16 \times 16$, resulting in 196 patch tokens per image. A learnable [CLS] token is prepended to the sequence of patch embeddings, and the combined sequence is passed through the transformer encoder. The final layer output corresponding to the [CLS] token serves as the global representation (a 768-dimensional vector), while the patch embeddings from the same layer constitute the local representation, denoted as $f \in \mathbb{R}^{768 \times 196}$. The embedding dimension of ViT-B/16 is fixed at 768.\\
\\
\textbf{Text Encoder.} We employ BioClinicalBERT \cite{me:biobert} as the text encoder, leveraging its implementation from the Hugging Face Transformers library \cite{transformer-library}. The model is initialized with weights pre-trained on the MIMIC-CXR corpus to ensure domain relevance. For each input report, we extract the hidden states from the final transformer layer as token-level (local) representations. To obtain sentence-level features, we apply a self-attention pooling mechanism across the tokens within each sentence, resulting in $\text{t}{i}^{\text{l}} \in \mathbb{R}^{P{i} \times D}$, where $P_{i}$ denotes the number of sentences in the $i$-th report. A second self-attention pooling layer is then applied across all sentence-level embeddings to generate a global report representation.

To enable joint modeling across modalities, we project both local and global representations from the image and text encoders into a shared embedding space. This is achieved by applying four independent linear projection layers, mapping each representation to a 128-dimensional embedding vector.

\subsection{Data Split of Downstream tasks}
Details of the dataset partitioning into training, validation, and test sets are provided in Table \ref{data}. For all downstream evaluations, models are trained using varying proportions of the training set, specifically 1\%, 10\%, and 100\%. All experiments are conducted on a workstation equipped with an NVIDIA RTX 4090 GPU to ensure computational efficiency and consistency.

\subsection{Implementation Details of Downstream Tasks}
\textbf{Semantic Segmentation.} We evaluate the segmentation capabilities of the proposed FaNe framework on two benchmark datasets: the RSNA Pneumonia Detection Challenge dataset \cite{rsna} and the SIIM-ACR Pneumothorax Segmentation dataset \cite{siim}. Our preprocessing pipeline closely follows the protocol established in MGCA \cite{rw:MGCA}. For the RSNA dataset, pneumonia masks are generated by converting the annotated bounding boxes into binary segmentation masks. Both input images and corresponding masks are resized to a spatial resolution of $224 \times 224$. To enhance training diversity, we apply data augmentation using the ShiftScaleRotate transformation from the albumentations library\footnote{\url{https://albumentations.ai/}}, which introduces random affine transformations including translation, scaling, and rotation. The augmentation parameters are set as follows: a maximum rotation of 10 degrees, a scale variation of up to 10\%, and an application probability of 0.5. Following augmentation, all images are normalized to the [0, 1] range prior to being fed into the segmentation network.

To assess the effectiveness of the pre-trained ResNet-50 backbone in downstream semantic segmentation, we adopt the U-Net architecture \cite{unet} with a ResNet-50 encoder, implemented via the Segmentation Models PyTorch library\footnote{\url{https://github.com/qubvel-org/segmentation_models.pytorch}}. Model training is conducted using the AdamW optimizer, configured with a learning rate of 5e-4 and a weight decay of 1e-6. In line with the strategy employed in GLoRIA \cite{bg:gloria}, we optimize the model using a composite loss function defined as $\alpha \times \text{FocalLoss} + \text{DiceLoss}$, where the weighting coefficient $\alpha$ is empirically set to 10 to balance class imbalance and spatial accuracy. The model is fine-tuned for a maximum of 50 epochs, with early stopping triggered if the validation loss fails to improve over 10 consecutive runs. The checkpoint corresponding to the lowest validation loss is retained for final evaluation on the test set.\\
\begin{table}[]
\centering
\begin{tabular}{ccccc}
\toprule
Task & Dataset & Train & Valid & Test \\ \midrule
\multirow{3}{*}{\begin{tabular}[c]{@{}c@{}}Linear\\ Classification\end{tabular}} & CheXpert & 186027 & 5000 & 202 \\
 & RSNA & 16010 & 5337 & 5337 \\
 & COVIDx & 23988 & 5998 & 400 \\ \midrule
\multirow{2}{*}{\begin{tabular}[c]{@{}c@{}}Semantic\\ Segmentation\end{tabular}} & RSNA & 16010 & 5337 & 5337 \\
 & SIIM & 8433 & 1807 & 1807 \\ \midrule
\multirow{2}{*}{\begin{tabular}[c]{@{}c@{}}Object\\ Detection\end{tabular}} & RSNA & 16010 & 5337 & 5337 \\
 & Object CXR & 6400 & 1600 & 1000 \\ \bottomrule
\end{tabular}
\caption{Details of data split.}
\label{data}
\end{table}
\\
\textbf{Object Detection.} we adopt a consistent and controlled setup. Specifically, when fine-tuning on the RSNA and Object-CXR datasets, only 1\% of the available training data is utilized, with a batch size of 8. For all other detection tasks, we increase the batch size to 16. No data augmentation techniques are applied to preserve the integrity of the original annotations. Each image is resized to a resolution of $224 \times 224$ and normalized to the [0, 1] range before being input into the detection network.

Following the optimization strategy in MGCA \cite{rw:MGCA}, we use the AdamW optimizer with a fixed learning rate of 5e-4 and a weight decay of 1e-6. No learning rate scheduling is applied during training. In our implementation, the standard Darknet-53 backbone is replaced with our pre-trained ResNet-50 encoder. Prior to training, the image encoder is frozen, and the remaining layers are randomly initialized. For object localization, we extract multi-scale feature representations from the 2nd, 3rd, and 4th bottleneck stages of the ResNet-50 backbone. Anchor boxes are adapted from the original YOLOv3 configuration \cite{yolov3}, with appropriate rescaling to match the $224 \times 224$ input size.

The model is fine-tuned for up to 50 epochs, and early stopping is triggered if no improvement in validation loss is observed over 10 consecutive runs. The model checkpoint with the lowest validation loss is preserved for final evaluation on the test set.
\\
\\
\textbf{Classification.} For fine-tuning on the CheXpert dataset, we adopt a batch size of 96 to accommodate its scale. In contrast, all other linear classification tasks are conducted using a batch size of 48. Image preprocessing mirrors the pipeline used for MIMIC-CXR: the longer side of each image is resized to 256 pixels, while the shorter side is padded with zeros to produce a square input of size $256 \times 256$. During training, we apply random cropping to obtain $224 \times 224$ patches, whereas centered cropping is used during validation and testing. All cropped images are subsequently normalized to the [0, 1] range before being passed to the classification model.

In the linear evaluation protocol, the pre-trained image encoder—either ResNet-50 or ViT-B/16—is frozen, and only the classification head is trained. This head is initialized with random weights. Optimization is performed using AdamW with a learning rate of 5e-4 and a weight decay of 1e-6. Each model is fine-tuned for up to 50 epochs, with early stopping triggered if the validation loss does not improve for 10 consecutive runs. The checkpoint corresponding to the lowest validation loss is retained for final evaluation on the test set.

\begin{figure}
    \centering
    \includegraphics[width=\linewidth]{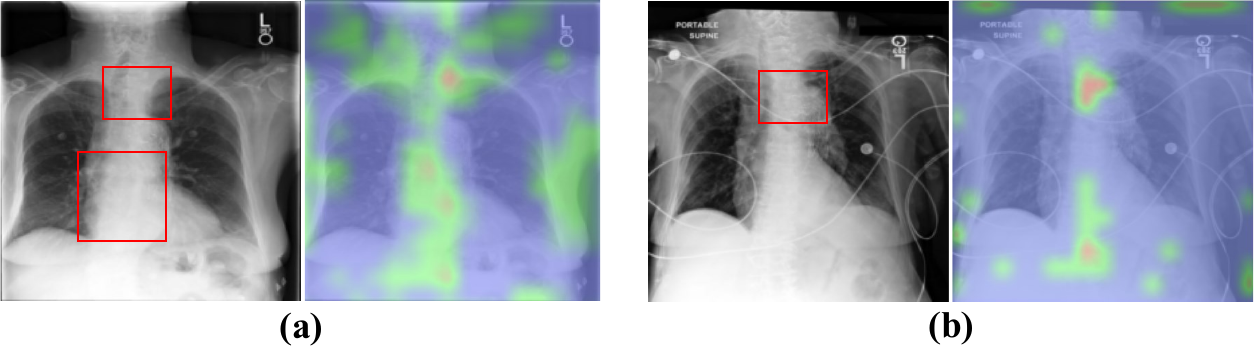}
    \caption{Representative cross-modality attention maps by ResNet-50. (a) The related sentence is "The \textbf{heart} size remains mildly enlarged and status post aortic stent graft repair of a descending \textbf{thoracic aortic} dissection." (b) The related sentence is "There is stable minimal widening of the ascending \textbf{aorta}."}
    \label{fig:attention}
\end{figure}
\subsection{Visualizations Using ResNet}
In addition to the visualization results based on the ViT backbone, we further evaluate the interpretability and effectiveness of our text-conditioned fine-grained alignment using a ResNet-50 encoder. As illustrated in Figure~\ref{fig:attention}, the attention maps reveal image regions with high cross-modal relevance, where brighter pixels correspond to higher attention weights. Our sparse attention pooling mechanism, when coupled with ResNet-50, successfully localizes clinically meaningful regions such as the heart and aorta, conditioned on the associated textual descriptions.

However, compared to ViT-based visualizations, those generated using ResNet-50 exhibit greater noise. We attribute this to the overlapping receptive fields inherent in convolutional architectures, which may hinder precise region-to-phrase alignment and lead to erroneous positive pair associations. In contrast, ViT processes the image as a sequence of non-overlapping patches, offering improved granularity that facilitates more accurate fine-grained alignment between visual and textual modalities.

\end{document}